\documentclass[letterpaper]{article} 
\usepackage[preprint]{aaai2027}  
\usepackage[hyphens]{url}  
\usepackage{graphicx} 
\usepackage{natbib}  
\usepackage{caption} 
\usepackage{algorithm}
\usepackage{algorithmic}
\usepackage{amsmath}
\usepackage{amssymb}
\usepackage{colortbl}
\usepackage{booktabs}
\usepackage{needspace}
\usepackage{pgfplots}
\usepgfplotslibrary{groupplots}
\pgfplotsset{compat=1.18}

\definecolor{hdrbg}{HTML}{DCE6F1}   
\definecolor{oursbg}{HTML}{FFF2CC}  
\definecolor{curveblue}{HTML}{356A9A}
\definecolor{defaultred}{HTML}{C84B4B}
\definecolor{baselinegray}{HTML}{767676}

\title{Sparse MLLM Anchors, Dense Adaptation: Breaking the Self-Referential Loop in Wild Test-Time Adaptation}
\author{\fontsize{12}{14}\selectfont\mbox{Zhenbin Wang, Lei Zhang$^{*}$, Lituan Wang, Yan Wang, Zhao Zhang, Wei Huang}}
\affiliations{Sichuan University\\\texttt{wangzhenbin@stu.scu.edu.cn}}

\begin{document}

\maketitle
\begingroup
\renewcommand{\thefootnote}{\fnsymbol{footnote}}
\footnotetext[1]{The corresponding author
}
\endgroup

\begin{abstract}
Wild test-time adaptation (WTTA) updates a source model online under small test batches, concurrent distribution shifts, and time-varying class imbalance. Most WTTA methods derive their adaptation signals, including predictive uncertainty, sample reliability, and local feature geometry, from the model being adapted. When the source model is unreliable under shift, these signals can reinforce its own errors, forming a \textit{self-referential loop}. We introduce MASA (\underline{M}ultimodal-LLM-\underline{A}nchored \underline{S}emantic \underline{A}daptation), which complements model-internal evidence with structured semantic descriptions from a frozen multimodal large language model (MLLM). To limit inference cost, MASA queries the MLLM only for a small set of diverse, reliability-ranked anchors. The resulting descriptions capture the object family and nuisance factors such as style, viewpoint, and occlusion. MASA encodes these descriptions, propagates them to neighboring test samples, and stores the resulting visual--semantic information in an online prototype memory. Descriptor-aware retrieval from this memory provides an auxiliary target for lightweight adaptation of normalization-affine parameters. We evaluate MASA on the WTTA ImageNet-C benchmark under limited-batch, mixed-domain, and imbalanced-label-shift settings with ResNet and ViT backbones. Code is available at
\leavevmode\pdfstartlink attr{/Border [0 0 0]}
user{/Subtype /Link /A << /S /URI /URI (https://github.com/wongzbb/MASA) >>}%
this link\pdfendlink.
\end{abstract}

\section{Introduction}
Deep models deployed in open-world environments must contend with test distributions that depart from the training distribution because of sensor noise, weather, style, viewpoint, and other changes. Test-time adaptation (TTA) updates a pretrained source model online using an unlabeled test stream, without labels or continued access to the source training set. WTTA considers a more demanding setting in which batches can contain a single sample, several shifts can coexist within one stream, and class frequencies can be imbalanced and time-varying. Each update therefore extract a useful signal from little and potentially unreliable evidence while remaining efficient enough for online inference.

\begin{figure}[t]
\centering
\includegraphics[width=\columnwidth]{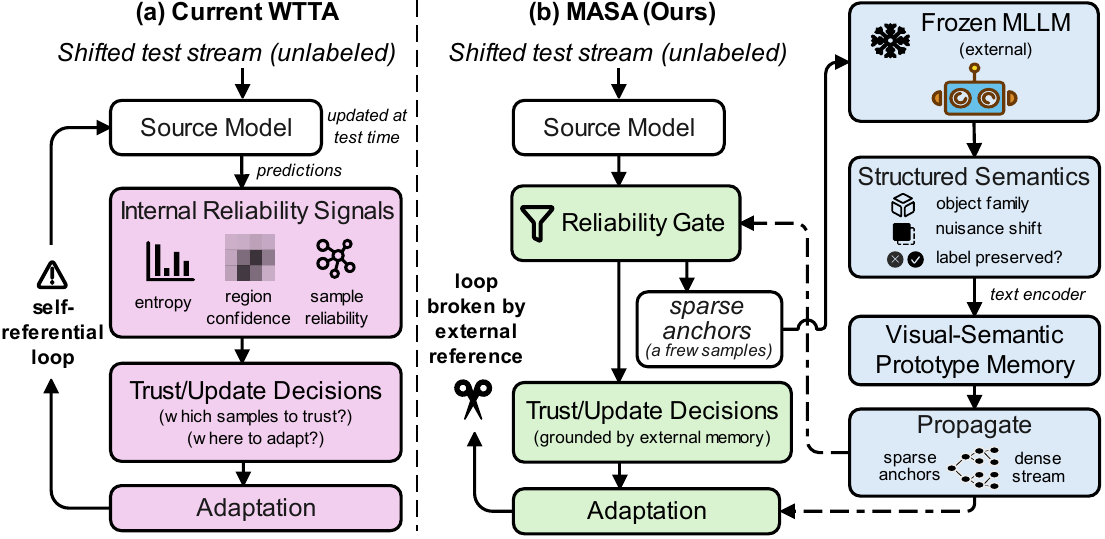}
\vspace{-1.5em}
\caption{\textbf{Comparison of adaptation evidence.} \textit{Left:} many WTTA methods estimate reliability and construct adaptation targets from the model being adapted, allowing prediction errors to influence subsequent updates. \textit{Right:} MASA queries a frozen MLLM on sparse anchors, propagates the encoded descriptors through local feature neighborhoods, and stores them in a visual--semantic prototype memory for descriptor-aware retrieval and memory admission.}
\vspace{-1.5em}
\label{fig:loop}
\end{figure}

Most WTTA methods improve the reliability of signals already produced by the source model. Entropy minimization \citep{tent} sharpens individual predictions. Sample-selection methods \citep{eata,sar,deyo} retain samples according to entropy, sharpness, or perturbation sensitivity, while normalization- and prototype-based methods stabilize online statistics or class centers. ReCAP \citep{recap} instead regularizes confidence and consistency within local feature neighborhoods. Although these methods act on different units, from individual predictions to selected samples and local regions, the evidence used to judge reliability remains largely derived from the model's own predictions and representations.

Reliance on internal evidence becomes risky when the model is itself unreliable under distribution shift. Prediction entropy, sharpness, perturbation sensitivity, and region confidence may then provide unreliable criteria for selecting samples or updating representations. We call this coupling a \emph{self-referential loop}: an incorrect but confident prediction can be selected for adaptation and subsequently reinforce the same error (Fig.~\ref{fig:loop}). This failure can be particularly consequential in small batches and imbalanced streams, where a few updates have disproportionate influence. Generative priors introduce external information by restoring inputs toward the source manifold \citep{dda} or distilling a diffusion score \citep{dusa}, but they act at the pixel or score level and may add substantial per-sample computation. They also do not explicitly describe the nuisance factors affecting an observation or whether recognizable object evidence remains under the degradation.

To address this limitation, we propose MASA, which introduces semantic evidence about image content and nuisance factors into the adaptation process. A frozen MLLM can produce such descriptions without conditioning on the classifier's logits, complementing signals internal to the source model. Recent work has shown the value of MLLM-based reasoning in out-of-distribution (OOD) detection \citep{ants}, but WTTA imposes different constraints. Querying every test sample is expensive, observations arrive online, and the source backbone need not share an image--text embedding space. The practical problem is therefore to convert a small number of MLLM descriptions into a persistent signal that can guide adaptation throughout the stream.

MASA converts sparse MLLM observations into dense adaptation guidance. It queries the frozen MLLM only for a small set of diverse, reliability-ranked anchors and requests descriptions of the object family and nuisance factors affecting each image. These responses serve neither as pseudo-labels nor as direct class predictions. MASA instead encodes them as semantic descriptors, propagates descriptor information across local feature neighborhoods, and integrates it into an online visual--semantic prototype memory. During adaptation, semantic agreement informs prototype retrieval, and the retrieved prototypes provide an auxiliary consistency target. Reusing each description across nearby samples amortizes MLLM queries over the stream. The framework requires no image--text alignment in the source classifier and updates only selected normalization-affine parameters.

Our contributions are threefold. First, we formulate a sparse semantic anchoring strategy that complements source-model reliability signals with structured descriptions from a frozen MLLM, without requiring a text-aligned source classifier. Second, we introduce a visual--semantic prototype memory that propagates anchor descriptors through local feature neighborhoods and uses descriptor-aware retrieval to regularize normalization-affine adaptation. Third, extensive experiments on ImageNet-C demonstrate that MASA achieves state-of-the-art performance across limited-batch, mixed-domain, and time-varying label-shift protocols with both ResNet and ViT backbones.

\section{Method}
\label{sec:method}

\subsection{Problem Setup and Overview}
Let $f_{\theta^{\mathrm{src}}}$ denote a $C$-class source classifier that outputs logits. We initialize $\theta_0=\theta^{\mathrm{src}}$ and receive an unlabeled batch $X_t=\{x_{t,i}\}_{i=1}^{B_t}$, with $B_t=|X_t|$, at each test step $t$. Labels are unavailable during adaptation and serve only to evaluate stream accuracy. Under MASA's predict-then-adapt protocol, the pre-update model $\theta_t$ predicts $X_t$ before its update affects subsequent batches.

MASA makes one stream pass and updates only affine scale and shift parameters in selected normalization layers; the classifier head and all other parameters remain fixed. Before $X_t$, the model and persistent states contain only earlier batches. The observation-level FIFO candidate window $\mathcal{W}_t$, anchor bank $\mathcal{A}_t$, and bounded prototype memory $\mathcal{M}_t$ are initially empty; bounded transition and coverage histories and a scalar recovery statistic are also maintained. We omit $t$ from sample- and cluster-level quantities when unambiguous.

\begin{figure*}[t]
\centering
\includegraphics[width=\textwidth]{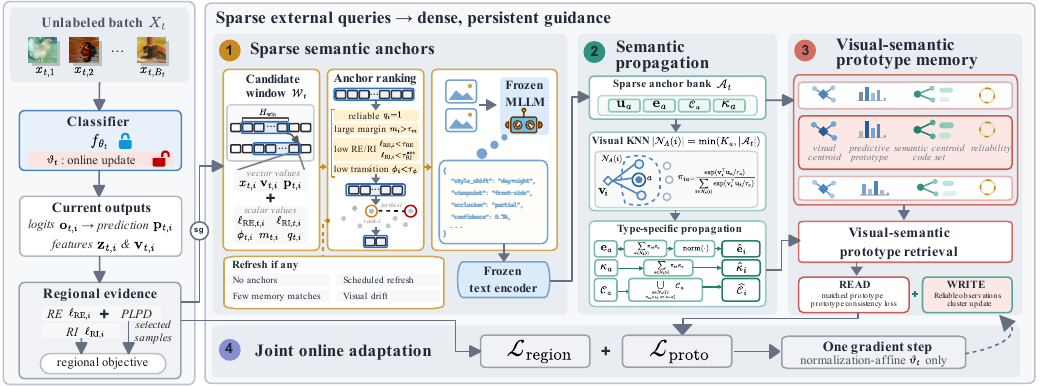}
\vspace{-1.5em}
\caption{\textbf{MASA overview.} Gray modules form the regional-evidence path. MASA ranks buffered observations, selects diverse anchors, and queries a frozen MLLM for object and nuisance descriptors. It propagates these descriptors to nearby samples and stores them in a bounded visual--semantic memory for descriptor-aware retrieval. Retrieved prototypes provide an auxiliary consistency target for normalization-affine adaptation.}
\label{fig:framework}
\vspace{-1em}
\end{figure*}

MASA complements regional evidence with reusable semantic observations (Fig.~\ref{fig:framework}). It ranks buffered observations by regional statistics, queries a frozen MLLM on a small, visually diverse subset, and propagates the descriptors to nearby samples. A bounded visual--semantic memory consolidates them into prototype targets that regularize the regional update.

\Needspace{6\baselineskip}
\subsection{Preliminaries and Two-Stage Evidence Selection}

MASA combines ReCAP regional confidence~\citep{recap} and DeYO patch sensitivity~\citep{deyo} as an ordered filter for regional adaptation, then reuses their detached statistics to rank semantic queries.

\paragraph{ReCAP regional confidence.}
For sample $i$, let $\mathbf{z}_i\in\mathbb{R}^{d_v}$ be the pre-update, unnormalized feature immediately before the final linear classifier, where $d_v$ is its dimension. The fixed head has weight $\mathbf{W}\in\mathbb{R}^{C\times d_v}$ and bias $\mathbf{b}\in\mathbb{R}^{C}$; $\mathbf{w}_c^\top$ denotes row $c$ of $\mathbf{W}$.
ReCAP's sampling-free proxy uses a fixed source-side variance vector $\boldsymbol{\sigma}_{z,\mathrm{src}}^2\in\mathbb{R}_{\geq0}^{d_v}$, whose $j$-th entry $[\boldsymbol{\sigma}_{z,\mathrm{src}}^2]_j=\operatorname{Var}_{\mathrm{src}}[z_j]$ is the variance of coordinate $j$ in the same feature space. This model-specific statistic is provided once and fixed throughout the target stream.

An effective regional coefficient $\rho_{\mathrm{reg}}>0$ scales this statistic. The resulting diagonal scale $\boldsymbol{\Lambda}_{\mathrm{reg}}$ induces a class-pair moment matrix $\mathbf{M}\in\mathbb{R}_{>0}^{C\times C}$:
\begin{equation}
\begin{aligned}
\boldsymbol{\Lambda}_{\mathrm{reg}}
&=\rho_{\mathrm{reg}}\operatorname{Diag}\!\left(
\boldsymbol{\sigma}_{z,\mathrm{src}}^2\right),\\
M_{cc'}&=\exp\!\left[
\frac{1}{2}(\mathbf{w}_c-\mathbf{w}_{c'})^\top
\boldsymbol{\Lambda}_{\mathrm{reg}}
(\mathbf{w}_c-\mathbf{w}_{c'})\right].
\end{aligned}
\label{eq:region-factors}
\end{equation}
The fixed head gives logits $\mathbf{o}_i=\mathbf{W}\mathbf{z}_i+\mathbf{b}\in\mathbb{R}^{C}$ and probabilities $\mathbf{p}_i=\operatorname{softmax}(\mathbf{o}_i)$ in the $C$-class simplex; $o_{i,c}$ and $p_{i,c}$ denote their $c$-th entries.
The inherited class-wise logit correction is $\boldsymbol{\delta}^{\mathrm{reg}}=(\delta_1^{\mathrm{reg}},\ldots,\delta_C^{\mathrm{reg}})^\top$, with $\delta_c^{\mathrm{reg}}=\lambda_{\delta}\lVert\mathbf{w}_c\rVert_2^2$, where $\lambda_{\delta}>0$ is the correction coefficient.
Both $\mathbf{M}$ and $\boldsymbol{\delta}^{\mathrm{reg}}$ are computed once and fixed.

For sample $i$, the moment matrix transforms $\mathbf{p}_i$ into the unnormalized regional mass $\boldsymbol{\nu}_i\in\mathbb{R}_{>0}^{C}$, while the corrected logits define a second distribution $\breve{\mathbf{p}}_i$:
\begin{equation}
\begin{aligned}
\nu_{i,c}&=\sum_{c'=1}^{C}p_{i,c'}M_{cc'}, \quad
\breve{\mathbf{p}}_i
=\operatorname{softmax}\!\left(
\mathbf{o}_i+\boldsymbol{\delta}^{\mathrm{reg}}\right).
\end{aligned}
\end{equation}
ReCAP combines these quantities into the closed-form Regional Entropy (RE) and Regional Instability (RI) proxies:
\begin{equation}
\begin{aligned}
\ell_{\mathrm{RE},i}
&=\sum_{c=1}^{C}\breve p_{i,c}
  \log\frac{\nu_{i,c}}{p_{i,c}}, \quad
\ell_{\mathrm{RI},i}
=\sum_{c=1}^{C}p_{i,c}\log\nu_{i,c}.
\end{aligned}
\label{eq:region-terms}
\end{equation}
RE measures regional predictive uncertainty, whereas RI captures within-region variation; lower values indicate more reliable evidence.

\paragraph{DeYO patch sensitivity.}
Following DeYO, we test reliance on coherent spatial structure by resizing an image to dimensions divisible by $P$, partitioning it into a $P\times P$ grid, and randomly permuting the cells. Let $x_i^{\mathrm{shuf}}$ be this view and $\mathbf{p}_i^{\mathrm{shuf}}=\operatorname{softmax}(f_{\theta_t}(x_i^{\mathrm{shuf}}))$ its no-gradient prediction. For the original pseudo-label $\widehat y_i$, the pseudo-label probability difference (PLPD) is
\vspace{-0.5em}\begin{equation}
\begin{aligned}
\widehat y_i&=\arg\max_c p_{i,c},\quad
\Delta_i^{\mathrm{PLPD}}
=p_{i,\widehat y_i}
  -p^{\mathrm{shuf}}_{i,\widehat y_i}.
\end{aligned}
\vspace{-0.5em}
\end{equation}
A large $\Delta_i^{\mathrm{PLPD}}$ means spatial disruption weakens the original pseudo-label, indicating reliance on coherent image content.

\paragraph{Two-stage selection and regional objective.}
To avoid unnecessary shuffled passes, MASA applies the cheaper RE test before PLPD, yielding the adaptation set
\begin{equation}
\mathcal{S}_t=\left\{i:\ell_{\mathrm{RE},i}<\tau_{\mathrm{RE}},
 \ \Delta_i^{\mathrm{PLPD}}>\tau_{\mathrm{PLPD}}\right\}.
\label{eq:region-selection}
\end{equation}
Here, $\tau_{\mathrm{RE}}$ and $\tau_{\mathrm{PLPD}}$ are the respective acceptance thresholds. The set $\mathcal{S}_t$ determines which samples contribute to the regional objective. A separate reliability gate provides an additional signal for anchor ranking, prototype adaptation, and memory admission.

On retained samples, MASA optimizes reliability-weighted regional terms. Let $\operatorname{sg}[\cdot]$ denote stop-gradient; the RE threshold is also the reweighting reference, $\omega_{\max}^{\mathrm{reg}}$ caps sample weights, and $\lambda_{\mathrm{RI}}$ balances the proxies. The weights and batch objective are
\begin{equation}
\begin{aligned}
\omega_i^{\mathrm{reg}}&=\min\!\left\{
 \exp\!\big(\tau_{\mathrm{RE}}-\operatorname{sg}[\ell_{\mathrm{RE},i}]\big),
 \omega_{\max}^{\mathrm{reg}}\right\},\\
\mathcal{L}_{\mathrm{region}}
&=\frac{1}{|\mathcal{S}_t|}\sum_{i\in\mathcal{S}_t}
\omega_i^{\mathrm{reg}}
\big(\ell_{\mathrm{RE},i}+\lambda_{\mathrm{RI}}\ell_{\mathrm{RI},i}\big).
\end{aligned}
\vspace{-0.5em}
\label{eq:region-loss}
\end{equation}
The detached weight favors lower-RE samples, while $\omega_{\max}^{\mathrm{reg}}$ limits individual influence. When $\mathcal{S}_t$ is empty, $\mathcal{L}_{\mathrm{region}}$ is a differentiable zero.

\Needspace{6\baselineskip}
\subsection{Reliability-Guided Semantic Anchors}

Querying every observation is costly and redundant. MASA instead refreshes a small, reliability-ranked, visually diverse anchor set at initialization, periodically, or when drift and retrieval coverage indicate stale semantics.

We normalize the pre-head feature as $\mathbf{v}_{t,i}=\mathbf{z}_{t,i}/\lVert\mathbf{z}_{t,i}\rVert_2$, making dot products cosine similarities. Hence, $\mathbf{v}_{t,i}\in\mathbb{R}^{d_v}$ and $\lVert\mathbf{v}_{t,i}\rVert_2=1$. Let $p_{t,i,(1)}\geq p_{t,i,(2)}$ be the two largest entries of $\mathbf{p}_{t,i}$; their margin $m_{t,i}=p_{t,i,(1)}-p_{t,i,(2)}$ measures top-1 confidence. For short-term stability, $\mathcal{H}^{-}_{t,i}=(y^-_{t,i,1},\ldots,y^-_{t,i,L_{t,i}})$ contains the top-1 predictions immediately preceding $x_{t,i}$ in loader order, where $L_{t,i}\leq H_{\mathrm{hist}}$.
Let $\mathbf{1}[\cdot]$ denote the indicator function. The corresponding label-transition rate is
\begin{equation}
\phi_{t,i}=
\begin{cases}
\displaystyle\frac{1}{L_{t,i}-1}\sum_{j=2}^{L_{t,i}}
\mathbf{1}[y^-_{t,i,j}\neq y^-_{t,i,j-1}],
&L_{t,i}\geq2,\\[5pt]
0,&L_{t,i}<2.
\end{cases}
\label{eq:transition-rate}
\end{equation}
Thus, $\phi_{t,i}$ is the recent rate of adjacent prediction changes, not a repeated-image history. Combining it with RE, RI, and the probability margin, we reuse $\tau_{\mathrm{RE}}$ and introduce thresholds $\tau_{\mathrm{RI}}^{\mathrm{anc}}$, $\tau_m$, and $\tau_\phi$ for RI, margin, and transition rate, respectively, giving
\begin{equation}
q_i=\mathbf{1}\!\left[
\begin{aligned}
\ell_{\mathrm{RE},i}<\tau_{\mathrm{RE}}
&{}\land\ \ell_{\mathrm{RI},i}<\tau_{\mathrm{RI}}^{\mathrm{anc}}\\
&{}\land\ m_i>\tau_m
\ \land\ \phi_i<\tau_{\phi}
\end{aligned}
\right].
\label{eq:anchor-reliability}
\end{equation}
The gate $q_i=1$ requires low regional uncertainty and instability, a large margin, and stable predictions; it is reused for anchor ranking, prototype adaptation, and memory admission.

Each arriving sample contributes the detached record $\mathsf{r}_{t,i}:=(x_{t,i},\mathbf{v}_{t,i},\mathbf{p}_{t,i},\ell_{\mathrm{RE},t,i},\ell_{\mathrm{RI},t,i},m_{t,i},\phi_{t,i},q_{t,i})$. Besides the RGB input $x_{t,i}\in\mathbb{R}^{3\times H_x\times W_x}$ with height $H_x$ and width $W_x$, the record stores a $d_v$-dimensional visual feature, a $C$-dimensional prediction, and the five scalars $\ell_{\mathrm{RE},t,i}$, $\ell_{\mathrm{RI},t,i}$, $m_{t,i}$, $\phi_{t,i}$, and $q_{t,i}$, all fixed at arrival. With $\mathbin{\Vert}$ denoting sequence concatenation and $\operatorname{tail}_H$ retaining the latest $H$ records, the current batch gives
\begin{equation}
\widetilde{\mathcal{W}}_t=\operatorname{tail}_{H_{\mathrm{win}}}\!\left(
\mathcal{W}_t{\mathbin{\Vert}}(\mathsf{r}_{t,1},\ldots,\mathsf{r}_{t,B_t})\right),
\quad |\widetilde{\mathcal{W}}_t|{\leq} H_{\mathrm{win}}.
\end{equation}
Excluding images, the $N_t=|\widetilde{\mathcal{W}}_t|\leq H_{\mathrm{win}}$ records form an $N_t\times(d_v+C+5)$ metadata array. Since $H_{\mathrm{win}}$ counts observations, not steps, its temporal span varies with $B_t$. Refresh reads without removing records, and absent recovery, $\mathcal{W}_{t+1}=\widetilde{\mathcal{W}}_t$. A \emph{window candidate} is a buffered observation eligible for querying, distinct from a candidate memory cluster.

At a refresh event, $R_i^{\mathrm{anc}}$ ranks buffered candidates by model-derived reliability:
\begin{align}
R_i^{\mathrm{anc}}{=}&\!\left(\!1{-}
 \min\!\left\{\frac{\ell_{\mathrm{RE},i}}{\log C},c_{\mathrm{clip}}\right\}\!\right)
{+}\left(\!1{-}
 \min\!\left\{\frac{\ell_{\mathrm{RI},i}}
 {\tau_{\mathrm{RI}}^{\mathrm{anc}}},c_{\mathrm{clip}}\right\}\!\right) \notag\\
&+m_i-\phi_i+q_i.
\label{eq:anchor-ranking}
\end{align}
Clipping prevents either regional proxy from dominating, while higher scores favor low RE and RI, a large margin, and few transitions. To avoid near duplicates, MASA shortlists the top $\min(|\widetilde{\mathcal{W}}_t|,r_{\mathrm{pool}}B_a)$ records and runs farthest-point sampling on $\mathbf{v}_i$, seeded by the highest-ranked record, to select at most $B_a$ anchors. Here, $B_a$ is the query budget, $r_{\mathrm{pool}}$ the shortlist multiplier, and $c_{\mathrm{clip}}$ the bound on each regional contribution.

Refresh uses recent feature-to-memory distance to detect stale coverage. Let $\{\boldsymbol{\mu}_k\}_{k=1}^{K_t}$ be the unit visual centroids of the $K_t\leq K_{\max}$ clusters, and let $\mathcal{R}_t$ index the $\min(H_D,|\widetilde{\mathcal{W}}_t|)$ most recent records of $\widetilde{\mathcal{W}}_t$, where $H_D$ is the drift horizon. Their mean nearest-centroid cosine distance is
\begin{equation}
D_t=\frac{1}{|\mathcal{R}_t|}\sum_{j\in\mathcal{R}_t}
\left(1-\max_{1\leq k\leq K_t}
\mathbf{v}_j^\top\boldsymbol{\mu}_k\right),
\end{equation}
where $D_t=1$ if either $\mathcal{R}_t$ or $\mathcal{M}_t$ is empty, favoring refresh without coverage. Larger $D_t$ indicates poorer memory coverage.

Feature drift misses failed descriptor matching. After completed step $s$, let $r_s^{\mathrm{cov}}$ be the fraction of all $B_s$ samples whose descriptor-aware retrieval score exceeds the assignment threshold $\tau_{\mathrm{assign}}$ defined below. Coverage retains the latest $H_{\mathrm{cov}}$ completed steps. We use the same numerical horizon for $H_{\mathrm{cov}}$ and $H_{\mathrm{win}}$, but the former counts steps and the latter observations. Once the history has $H_{\mathrm{cov}}$ entries, the average before step $t$ is
\begin{equation}
\bar r_t^{\mathrm{cov}}
=\frac{1}{H_{\mathrm{cov}}}
\sum_{s=t-H_{\mathrm{cov}}}^{t-1}r_s^{\mathrm{cov}}.
\end{equation}
MASA refreshes whenever $\mathcal{A}_t=\varnothing$, including after initialization or recovery, every $T_{\mathrm{ref}}$ adaptation steps, or when $D_t>\tau_D$ or $\bar r_t^{\mathrm{cov}}<\tau_{\mathrm{cov}}$. Coverage is tested only after its history is full; $\tau_D$ and $\tau_{\mathrm{cov}}$ are the drift and coverage thresholds.

\subsection{Semantic Grounding and Propagation}

Selected anchors provide visual coverage but not content or nuisance descriptions; MASA obtains these from a frozen MLLM and propagates them through visual neighborhoods.

For a selected record $\mathsf{r}_{s,j}\in\widetilde{\mathcal{W}}_t$, where $s\leq t$, assign anchor index $a$ and reuse its detached feature as $\mathbf{u}_a=\mathbf{v}_{s,j}$. The query contains only the image and a fixed, task-agnostic prompt, with neither classifier predictions nor source-class names. The frozen MLLM returns open-vocabulary descriptions of object family, scene, style shift, viewpoint, and occlusion, plus response confidence $c_a^{\mathrm{resp}}\in[0,1]$ and object recognizability $c_a^{\mathrm{obj}}\in[0,1]$, which estimates whether primary-object evidence survives the degradation. Descriptions occupy a semantic space separate from the classifier's $C$-way output, while the scores determine reliability.

\begin{figure}[t]
\centering
\includegraphics[width=0.88\columnwidth]{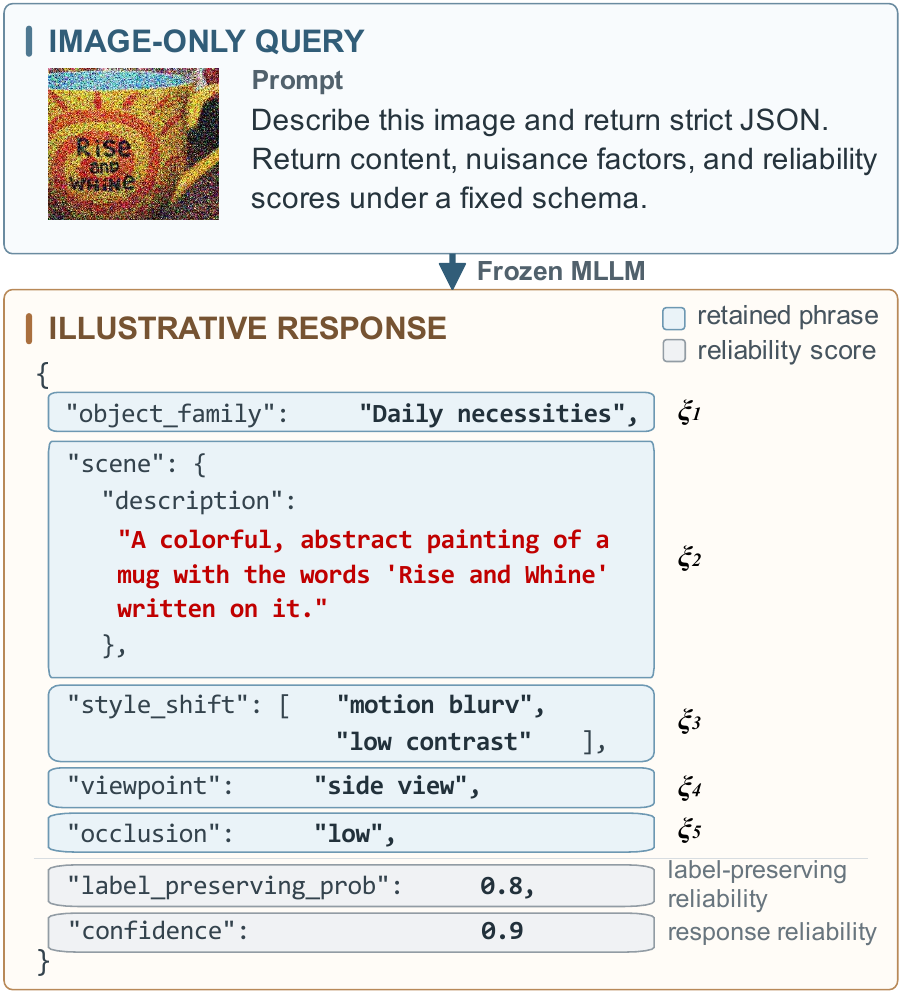}
\caption{\textbf{From an image-only MLLM query to descriptor phrases.} A fixed task-agnostic prompt returns structured fields. Each value is indexed by $l$ and denoted by $\xi_l$; list entries remain separate, while gray scores determine $\kappa_a$ and are not embedded.}
\label{fig:semantic-phrases}
\end{figure}

We discard empty and \texttt{unknown} values, serialize each retained value as \texttt{field: value}, and represent list elements separately. Denote the resulting $L$ phrases by $(\xi_1,\ldots,\xi_L)$ (Fig.~\ref{fig:semantic-phrases}). If none remains, the fallback \texttt{object\_family: unknown object} ensures $L\geq 1$. We use $\mathcal{C}_a$ for the retained code set, which excludes this fallback.

A frozen encoder $E_{\mathrm{text}}:\mathcal{T}\rightarrow\mathbb{R}^{d_s}$ maps phrase space $\mathcal{T}$ to a $d_s$-dimensional space distinct from classifier features. Let $\operatorname{norm}$ denote $\ell_2$ normalization. Uniform aggregation and descriptor reliability are
\begin{equation}
\begin{aligned}
\mathbf{e}_a&=\operatorname{norm}\!\left(
 \sum_{l=1}^{L}E_{\mathrm{text}}(\xi_l)\right),\\[-2pt]
\kappa_a&=\max\!\left(c_a^{\mathrm{resp}}c_a^{\mathrm{obj}},
 \kappa_{\min}\right).
\end{aligned}
\label{eq:anchor-semantics}
\end{equation}
Uniform aggregation avoids field weights, normalization gives unit descriptors in $\mathbb{R}^{d_s}$, and $\kappa_{\min}$ prevents zero reliability. The bank stores $(\mathbf{u}_a,\mathbf{e}_a,\mathcal{C}_a,\kappa_a)$, where $\lVert\mathbf{u}_a\rVert_2=1$ and $\mathcal{C}_a$ contains codes $\langle\textsf{field}{=}\textsf{value}\rangle$. For maximum neighborhood size $K_a$, its capacity is $\max(4K_{\max},K_a)$. Refresh appends tuples to $\mathcal{A}_t$ in query order and evicts the oldest when full; $\kappa_a$ controls the anchor's memory update.

Because few observations are queried, MASA estimates each descriptor from visual neighbors. For feature $\mathbf{v}_i$ and a nonempty bank, let $\mathcal{N}_{\!A}(i)$ contain the $\min(K_a,|\mathcal{A}_t|)$ anchors with largest $\mathbf{v}_i^\top\mathbf{u}_a$, where $K_a$ is the maximum neighborhood size. The nearest index is $a_i^\star\in\operatorname*{arg\,max}_{a\in\mathcal{N}_{\!A}(i)}\mathbf{v}_i^\top\mathbf{u}_a$. With temperature $\tau_a$, propagation is
\setlength{\jot}{1pt}
\begin{equation}
\begin{aligned}
\pi_{ia}
{=}\frac{\exp(\mathbf{v}_i^\top\mathbf{u}_a/\tau_a)}
{\sum_{b\in\mathcal{N}_{\!A}(i)}
 \exp(\mathbf{v}_i^\top\mathbf{u}_b/\tau_a)}, \quad \widehat\kappa_i
 {=}\!\!\!\!\sum_{a\in\mathcal{N}_{\!A}(i)}\pi_{ia}\kappa_a,\\
\widehat{\mathbf{e}}_i
{=}\operatorname{norm}\!\left(\!
 \sum_{a\in\mathcal{N}_{\!A}(i)}\pi_{ia}\mathbf{e}_a \!\right)\!, \quad
\widehat{\mathcal{C}}_i
{=}\!\!\!\!\!\!\bigcup_{\substack{a\in\mathcal{N}_{\!A}(i):\\
 \pi_{ia}>\epsilon_{\mathcal C}\ \mathrm{or}\ a=a_i^\star}} \!\!\!\!\!\!
 \mathcal{C}_a.
\end{aligned}
\label{eq:semantic-propagation}
\end{equation}
Here, $\epsilon_{\mathcal C}$ is the code-inclusion threshold. The weights $\pi_{ia}$ interpolate embeddings and reliabilities, producing $\widehat{\mathbf{e}}_i$ and $\widehat\kappa_i$; $\widehat{\mathcal{C}}_i$ collects codes from influential anchors and always considers the nearest, whose code set may be empty. If $\mathcal{A}_t$ is empty, $\widehat{\mathbf{e}}_i$ is absent, $\widehat\kappa_i=0$, and $\widehat{\mathcal{C}}_i=\varnothing$. All descriptor quantities are detached from the classifier graph.

\Needspace{6\baselineskip}
\subsection{Visual--Semantic Prototype Memory}

Propagation yields per-observation descriptors but no stream-level consolidation. MASA therefore maintains bounded visual, predictive, and semantic prototypes. Retrieval reads incoming memory; ordinary writes follow adaptation, except that refreshed anchors are upserted before retrieval to affect the current batch.

The memory contains $K_t\leq K_{\max}$ clusters, $\mathcal{M}_t=\{\mathfrak{m}_k\}_{k=1}^{K_t}$, each summarizing admitted observations and anchors through four descriptor states. The unit visual centroid is $\boldsymbol{\mu}_k\in\mathbb{R}^{d_v}$, and $\bar{\mathbf{p}}_k$ is the predictive prototype in the $C$-class simplex. The unit semantic centroid $\bar{\mathbf{e}}_k\in\mathbb{R}^{d_s}$ is absent when no semantic evidence exists. The set $\mathcal{C}_k$ accumulates codes $\langle\textsf{field}{=}\textsf{value}\rangle$ from $\widehat{\mathcal{C}}_i$ for observations or $\mathcal{C}_a$ for anchors.

Five statistics control trust and retention: reliability $\psi_k\in[0,1]$ averages write reliabilities, support $n_k\in\mathbb{N}_{+}$ counts admitted observations and anchors, $\mathrm{age}_k\in\mathbb{N}_0$ counts steps since the latest write, and drift $d_k\geq0$ averages incoming cosine distance. Finally, $\chi_k\in\{0,1\}$ distinguishes candidate ($\chi_k=0$) from committed ($\chi_k=1$) clusters. The state is
\begin{equation}
\mathfrak{m}_k=
\big(\boldsymbol{\mu}_k,\bar{\mathbf{p}}_k,
\bar{\mathbf{e}}_k,\mathcal{C}_k,
\psi_k,n_k,\mathrm{age}_k,d_k,\chi_k\big).
\end{equation}

\paragraph{Descriptor-aware retrieval.}
Retrieval seeks complementary evidence rather than visual proximity alone. It removes clusters below reliability $\tau_q$, giving
\begin{equation}
\mathcal{K}_t^{\mathrm{ret}}
=\{k\in\{1,\ldots,K_t\}:\psi_k\geq\tau_q\},
\end{equation}
where $\tau_q$ is the minimum reliability. Define normalized age $\widetilde{\mathrm{age}}_k=\min(\mathrm{age}_k/H_{\mathrm{win}},1)$ and Jaccard similarity $J(A,B)=|A\cap B|/|A\cup B|$, with $J(\varnothing,\varnothing)=0$. Let $s_{ik}^{\mathrm{sem}}=\cos(\widehat{\mathbf{e}}_i,\bar{\mathbf{e}}_k)$ when both vectors exist and $s_{ik}^{\mathrm{sem}}=0$ otherwise. Here $H_{\mathrm{win}}$ is reused only as a numerical cap for step-based age, despite denoting observation capacity above. All current-sample arguments are stop-gradient. For $k\in\mathcal{K}_t^{\mathrm{ret}}$,
\begin{equation}
\begin{aligned}
S_{ik}={}&\mathbf{v}_i^\top\boldsymbol{\mu}_k
 +\mathbf{p}_i^\top\bar{\mathbf{p}}_k
 +s_{ik}^{\mathrm{sem}}\\
&+J(\widehat{\mathcal{C}}_i,\mathcal{C}_k)
 -\alpha_{\mathrm{age}}\widetilde{\mathrm{age}}_k
 -\alpha_{\mathrm{cand}}(1-\chi_k).
\end{aligned}
\label{eq:retrieval-score}
\end{equation}
The positive terms in $S_{ik}$ measure visual, predictive, semantic-embedding, and code agreement; penalties weighted by $\alpha_{\mathrm{age}}$ and $\alpha_{\mathrm{cand}}$ discount stale and uncommitted clusters. If $\mathcal{K}_t^{\mathrm{ret}}$ is nonempty, set $k_i^\star=\arg\max_{k\in\mathcal{K}_t^{\mathrm{ret}}}S_{ik}$ and accept it when $S_{i k_i^\star}>\tau_{\mathrm{assign}}$, where $\tau_{\mathrm{assign}}$ is the assignment threshold. Other samples remain unassigned.

\paragraph{Prototype adaptation.}
An accepted cluster supplies fixed visual and predictive targets. The reliability gate blocks unreliable gradients, while a match weight grows with retrieval confidence at a rate controlled by $\tau_\omega$. Define
\begin{equation}
\begin{aligned}
\omega_i^{\mathrm{proto}}
&=\operatorname{sigmoid}\!\left(
 \frac{S_{i k_i^\star}-\tau_{\mathrm{assign}}}{\tau_\omega}\right),\\
\mathcal{I}_t&=\left\{i:k_i^\star\ \text{exists},\ 
 S_{i k_i^\star}>\tau_{\mathrm{assign}},\ q_i=1\right\}.
\end{aligned}
\label{eq:prototype-weight}
\end{equation}
Here $q_i$ is the gate in Eq.~\eqref{eq:anchor-reliability}. Let $d_{ik}^{\mathrm{sem}}=\operatorname{sg}[1-\cos(\widehat{\mathbf{e}}_i,\bar{\mathbf{e}}_k)]$ when both semantic vectors exist and $d_{ik}^{\mathrm{sem}}=0$ otherwise. For $i\in\mathcal{I}_t$, the discrepancy and batch objective are
\begin{equation}
\begin{aligned}
\ell_{\mathrm{proto},i}={}&
 \big(1-\mathbf{v}_i^\top\boldsymbol{\mu}_{k_i^\star}\big)\\
&+D_{\mathrm{KL}}\!\left(
 \bar{\mathbf{p}}_{k_i^\star}\,\|\,\mathbf{p}_i\right)
 +d_{i k_i^\star}^{\mathrm{sem}},\\
\mathcal{L}_{\mathrm{proto}}={}&
 \frac{1}{|\mathcal{I}_t|}
\sum_{i\in\mathcal{I}_t}
\omega_i^{\mathrm{proto}}\ell_{\mathrm{proto},i}.
\end{aligned}
\label{eq:prototype-objective}
\end{equation}
Here $D_{\mathrm{KL}}(\bar{\mathbf{p}}_k\|\mathbf{p}_i)=\sum_c\bar p_{k,c}\log(\bar p_{k,c}/p_{i,c})$ treats the stored prediction as target. Visual and predictive terms align the current output; semantics affects matching and $\omega_i^{\mathrm{proto}}$ through $S_{ik}$. Its detached discrepancy changes the recovery objective but not text gradients. All terms enter $\ell_{\mathrm{proto},i}$ with unit coefficients, and the loss is differentiable zero when $\mathcal{I}_t$ is empty.

\paragraph{Memory update.}
Memory updates are separated from gradient adaptation so that the current optimizer step cannot immediately rewrite its own targets. After adaptation, MASA writes detached features and predictions computed before the update. Let $\eta_0$ be the base update rate and let $\operatorname{clip}(x,0,1)$ truncate $x$ to $[0,1]$. The reliability gate and descriptor confidence jointly determine whether an observation may enter memory and, for a matched observation, how strongly it updates the cluster.

\Needspace{9\baselineskip}
\begin{equation}
\begin{aligned}
\rho_i^{\mathrm{mem}}
&=q_i\max(\widehat\kappa_i,\kappa_{\min}),\\
\eta_i^{\mathrm{mem}}&=\operatorname{clip}\!\left(
 \frac{\eta_0\rho_i^{\mathrm{mem}}}
 {\sqrt{n_{k_i^\star}+1}},0,1\right),
\qquad i\in\mathcal{I}_t.
\end{aligned}
\label{eq:memory-rate}
\end{equation}

The scalar $\rho_i^{\mathrm{mem}}$ is the admission reliability, and $\tau_{\mathrm{store}}$ is its minimum storage threshold. Samples with $\rho_i^{\mathrm{mem}}\leq\tau_{\mathrm{store}}$ are skipped. For an assignment accepted during batch retrieval, Eq.~\eqref{eq:memory-rate} updates $k=k_i^\star$. The factor $\sqrt{n_{k_i^\star}+1}$ makes well-supported clusters change more slowly.

An initially unassigned observation is compared again with the progressively updated memory. If this second retrieval accepts a match, we reuse $k_i^\star$ for that cluster and evaluate $\eta_i^{\mathrm{mem}}$ by the same rule in Eq.~\eqref{eq:memory-rate}, solely for the memory write. The observation is not added retrospectively to $\mathcal{I}_t$. If no match is accepted, it initializes a candidate cluster. Let $\operatorname{norm}_1$ denote $\ell_1$ normalization. With $+$ denoting the intermediate state after sample $i$, the three centroids are updated by
\begin{equation}
\begin{aligned}
\boldsymbol{\mu}_k^+&=\operatorname{norm}\!\left(
 (1-\eta_i^{\mathrm{mem}})\boldsymbol{\mu}_k
 +\eta_i^{\mathrm{mem}}\mathbf{v}_i\right),\\
\bar{\mathbf{p}}_k^+&=\operatorname{norm}_1\!\left(
 (1-\eta_i^{\mathrm{mem}})\bar{\mathbf{p}}_k
 +\eta_i^{\mathrm{mem}}\mathbf{p}_i\right),\\
\bar{\mathbf{e}}_k^+&=\operatorname{norm}\!\left(
 (1-\eta_i^{\mathrm{mem}})\bar{\mathbf{e}}_k
 +\eta_i^{\mathrm{mem}}\widehat{\mathbf{e}}_i\right).
\end{aligned}
\label{eq:centroid-update}
\end{equation}
If $\bar{\mathbf{e}}_k$ is absent, it is initialized by $\widehat{\mathbf{e}}_i$. If $\widehat{\mathbf{e}}_i$ is absent, the semantic update is skipped. Let $\lambda_{\mathrm{stat}}$ be the shared smoothing rate for cluster reliability and visual drift. The remaining matched-cluster statistics follow
\begin{equation}
\begin{aligned}
\mathcal{C}_k^+&=\mathcal{C}_k\cup\widehat{\mathcal{C}}_i,\quad
\psi_k^+=(1-\lambda_{\mathrm{stat}})\psi_k
 +\lambda_{\mathrm{stat}}\rho_i^{\mathrm{mem}},\\
d_k^+&=(1-\lambda_{\mathrm{stat}})d_k
 +\lambda_{\mathrm{stat}}\big(1-
 \boldsymbol{\mu}_k^\top\mathbf{v}_i\big),\\
n_k^+&=n_k+1,
\qquad \mathrm{age}_k^+=0.
\end{aligned}
\label{eq:memory-statistics}
\end{equation}
\Needspace{7\baselineskip}
Eq.~\eqref{eq:centroid-update} and~\eqref{eq:memory-statistics} are applied in loader order, so each $+$ state becomes the input to the next admitted write. An admitted observation with no accepted match initializes a candidate $k_{\mathrm{new}}$ by
\begin{equation}
\begin{aligned}
(\boldsymbol{\mu}_{k_{\mathrm{new}}},\bar{\mathbf{p}}_{k_{\mathrm{new}}},
 \bar{\mathbf{e}}_{k_{\mathrm{new}}},\mathcal{C}_{k_{\mathrm{new}}})
&=(\mathbf{v}_i,\mathbf{p}_i,
 \widehat{\mathbf{e}}_i,\widehat{\mathcal{C}}_i),\\
(\psi_{k_{\mathrm{new}}},n_{k_{\mathrm{new}}})
&=(\rho_i^{\mathrm{mem}},1),\\
(\mathrm{age}_{k_{\mathrm{new}}},d_{k_{\mathrm{new}}},
 \chi_{k_{\mathrm{new}}})&=(0,0,0).
\end{aligned}
\label{eq:cluster-init}
\end{equation}
If $\widehat{\mathbf{e}}_i$ is unavailable, the new cluster's semantic centroid is initialized as absent. The mandatory first anchor refresh bootstraps an initially empty memory. For each newly queried anchor, let $\rho_a^{\mathrm{anc}}=q_a\kappa_a$, where $\mathbf{p}_a$ and $q_a$ are taken from its window record and remain fixed after arrival. Anchor upserts do not use the sample-storage threshold $\tau_{\mathrm{store}}$. Instead, $\rho_a^{\mathrm{anc}}$ initializes or updates cluster reliability. We obtain $S_{ak}$ from Eq.~\eqref{eq:retrieval-score} by replacing $(\mathbf{v}_i,\mathbf{p}_i,\widehat{\mathbf{e}}_i,\widehat{\mathcal{C}}_i)$ with $(\mathbf{u}_a,\mathbf{p}_a,\mathbf{e}_a,\mathcal{C}_a)$.

If $\mathcal{K}_t^{\mathrm{ret}}$ is nonempty, let $k_a^\star=\arg\max_{k\in\mathcal{K}_t^{\mathrm{ret}}}S_{ak}$. When $S_{a k_a^\star}>\tau_{\mathrm{assign}}$, we set $\eta_a^{\mathrm{anc}}=\operatorname{clip}(\eta_0\rho_a^{\mathrm{anc}}/\sqrt{n_{k_a^\star}+1},0,1)$ and apply Eqs.~\eqref{eq:centroid-update}--\eqref{eq:memory-statistics} with $(\mathbf{v}_i,\mathbf{p}_i,\widehat{\mathbf{e}}_i,\widehat{\mathcal{C}}_i,\rho_i^{\mathrm{mem}},\eta_i^{\mathrm{mem}},k)$ replaced by $(\mathbf{u}_a,\mathbf{p}_a,\mathbf{e}_a,\mathcal{C}_a,\rho_a^{\mathrm{anc}},\eta_a^{\mathrm{anc}},k_a^\star)$. If the eligible set is empty or the score test fails, Eq.~\eqref{eq:cluster-init} applies to the anchor quantities. Promotion and capacity maintenance finish before batch retrieval, making retained anchor-seeded clusters immediately available.

\Needspace{9\baselineskip}
A new cluster remains a candidate until repeated support and sufficient reliability reduce the risk of storing an isolated error. It becomes committed when $n_k\geq n_{\min}$ and $\psi_k\geq\tau_q$, after which the commitment is irreversible. If $K_t>K_{\max}$, the memory retains the $K_{\max}$ clusters with the largest utility:
\begin{equation}
\begin{aligned}
U_k={}&\psi_k+\log(1+n_k)
-\widetilde{\mathrm{age}}_k-d_k+\chi_k.
\end{aligned}
\label{eq:memory-utility}
\end{equation}
The utility favors reliable, well-supported, committed clusters while discounting stale or drifting ones. Promotion and eviction are deferred until all precomputed assignments have been consumed, which keeps cluster indices stable during sequential writes. Maintenance uses the post-write, pre-increment ages. Each surviving age is then increased by one to form $\mathcal{M}_{t+1}$, so a cluster written at step $t$ has age one when step $t+1$ begins.

\vspace{-0.3em}
\subsection{Adaptation Objective and Online Procedure}

\begin{figure}[!t]
\centering
\includegraphics[width=0.93\columnwidth]{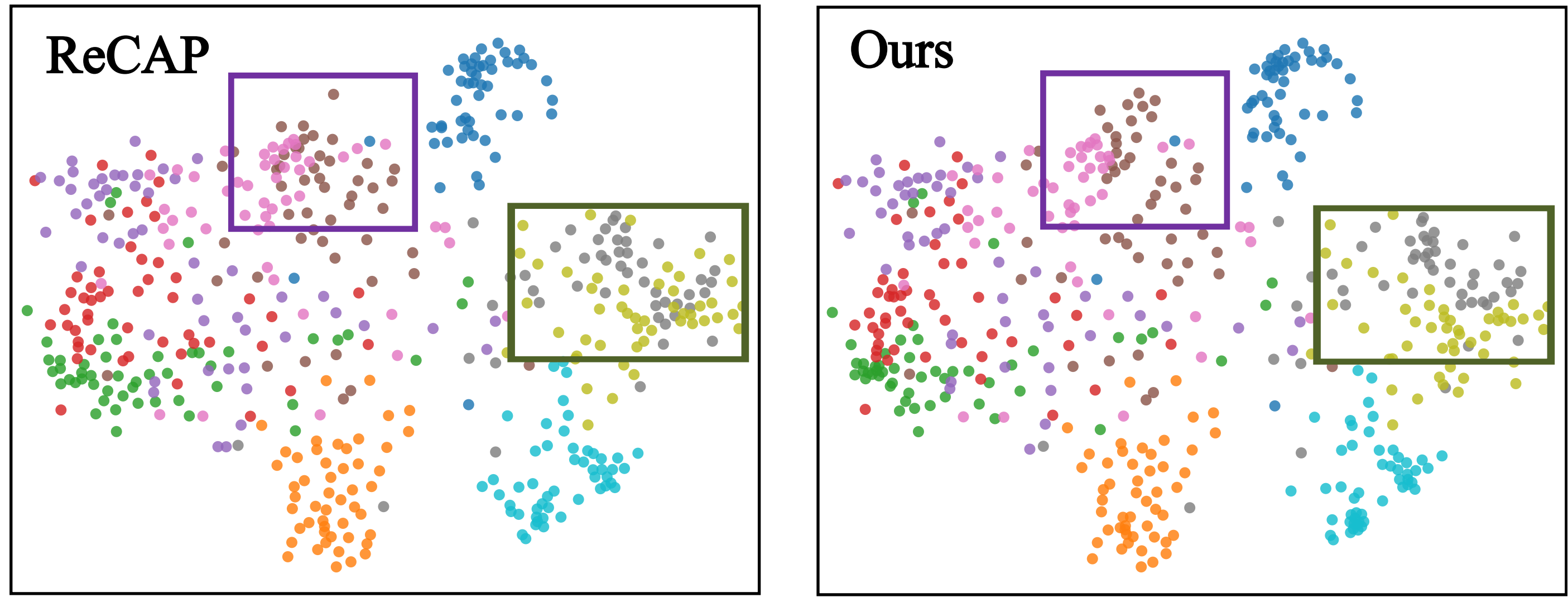}
\vspace{-0.6em}
\caption{\textbf{Feature-space visualization after online adaptation.}
t-SNE features of 10 classes produced by ReCAP and MASA on ImageNet-C under Elastic Transform at severity level 5.
Colors indicate classes, and boxes mark representative regions for comparison.}
\label{fig:feature-space}
\vspace{-1.2em}
\end{figure}

The regional and prototype losses address complementary failure modes: the former selects structurally reliable observations within the current batch, whereas the latter regularizes them against evidence accumulated across the stream. We next combine the two objectives and specify the causal ordering of prediction, adaptation, and state updates.

Let $\vartheta_t$ collect the trainable normalization-affine parameters of $\theta_t$.
For each test batch, MASA minimizes
\begin{equation}
\mathcal{L}_{\mathrm{MASA}}=
\mathcal{L}_{\mathrm{region}}
+\lambda_{\mathrm{proto}}\mathcal{L}_{\mathrm{proto}},
\label{eq:masa-objective}
\end{equation}
where $\lambda_{\mathrm{proto}}$ balances the two objectives. One gradient step maps $\vartheta_t$ to $\vartheta_{t+1}$. MASA returns the logits computed before this update, ensuring that the prediction for $X_t$ does not depend on its own adaptation step.

Following SAR \citep{sar}, MASA monitors an exponential moving average $\bar\ell_t$ of the adaptation objective and uses a small value as the recovery criterion.

\Needspace{6\baselineskip}
With momentum $\rho_{\mathrm{rec}}$, its update is
\begin{equation}
\bar\ell_t{=}
\begin{cases}
\mathcal{L}_{\mathrm{MASA}},&\text{if uninitialized},\\
\rho_{\mathrm{rec}}\bar\ell_{t-1}
{+}(1{-}\rho_{\mathrm{rec}})\mathcal{L}_{\mathrm{MASA}},&\text{otherwise}.
\end{cases}
\label{eq:recovery}
\end{equation}
Let $\tau_{\mathrm{rec}}$ be the recovery threshold. If $\bar\ell_t<\tau_{\mathrm{rec}}$, MASA restores the model and optimizer states saved at initialization. It also clears $\mathcal{W}_t$, $\mathcal{A}_t$, $\mathcal{M}_t$ and their coverage and transition histories, then marks the recovery average as uninitialized so that the next batch follows the first branch of Eq.~\eqref{eq:recovery}.

See the appendix for pseudocode of the complete predict-then-adapt procedure. Anchor refresh precedes retrieval so that new descriptions are immediately useful, whereas ordinary cluster writes use detached features and predictions computed before optimization. Unless recovery occurs, the post-append window becomes $\mathcal{W}_{t+1}$, and the final anchor and memory states are relabeled $(\mathcal{A}_{t+1},\mathcal{M}_{t+1})$.

\FloatBarrier

\section{Experiments}

\begin{table*}[!t]
\centering
\captionsetup{font=footnotesize,skip=1pt}
\setlength{\tabcolsep}{1.85pt}
\setlength{\aboverulesep}{0.35pt}
\setlength{\belowrulesep}{0.35pt}
\renewcommand{\arraystretch}{0.92}
\begin{minipage}[t]{0.486\textwidth}
\vspace{0pt}
\centering
\caption{Limited-batch evaluation on ImageNet-C at severity 5.}
\label{tab:bs1}
\tiny
\begin{tabular}{@{}l*{16}{c}@{}}
\toprule
  & \multicolumn{3}{c}{Noise}
  & \multicolumn{4}{c}{Blur}
  & \multicolumn{4}{c}{Weather}
  & \multicolumn{4}{c}{Digital}
  & \\
  \cmidrule(lr){2-4}
  \cmidrule(lr){5-8}
  \cmidrule(lr){9-12}
  \cmidrule(lr){13-16}
Method & Gau & Sho & Imp & Def & Gla & Mot & Zoo & Sno & Fro & Fog & Bri & Con & Ela & Pix & JPG & Avg \\
\midrule
\multicolumn{17}{c}{\itshape ResNet50-GN}\\[-1pt]
Source & 18.0&19.8&17.9&19.8&11.4&21.4&24.9&40.4&47.3&33.6&69.3&36.3&18.6&28.4&52.3&30.6 \\
MEMO & 18.5&20.5&18.4&17.1&12.6&21.8&26.9&40.4&47.0&34.4&69.5&36.5&19.2&32.1&53.3&31.2 \\
DDA & 42.4&43.3&42.3&16.6&19.6&21.9&26.0&35.7&40.1&13.7&61.2&25.2&37.5&46.6&54.1&35.1 \\
Tent & 2.5&2.9&2.5&13.5&3.6&18.6&17.6&15.3&23.0&1.4&70.4&42.2&6.2&49.2&53.8&21.5 \\
EATA & 24.9&28.0&25.8&18.3&17.0&31.2&29.8&42.5&44.1&41.3&70.9&44.2&27.6&46.8&55.4&36.5 \\
SAR & 25.5&28.0&24.9&18.7&16.3&28.6&31.4&46.2&44.9&33.4&72.8&44.3&15.3&47.1&56.1&35.6 \\
DeYO & 41.2&44.3&42.5&\underline{22.4}&24.7&41.8&21.9&\underline{54.8}&51.6&21.9&\underline{73.1}&53.2&\textbf{48.5}&59.8&59.6&44.1 \\
ReCAP & \underline{42.5}&\underline{44.4}&\underline{42.9}&19.4&\underline{25.0}&\underline{42.2}&\textbf{44.0}&49.7&\underline{52.4}&\underline{57.5}&72.9&\underline{53.6}&29.5&\underline{60.4}&\underline{60.0}&\underline{46.4} \\
\rowcolor{hdrbg}
\textbf{MASA} & \textbf{43.2}&\textbf{44.8}&\textbf{43.4}&\textbf{23.3}&\textbf{25.5}&\textbf{42.8}&\underline{42.9}&\textbf{55.5}&\textbf{52.6}&\textbf{58.3}&\textbf{73.2}&\textbf{54.0}&\underline{45.7}&\textbf{60.7}&\textbf{60.1}&\textbf{48.4} \\
\midrule
\multicolumn{17}{c}{\itshape ViT-Base-LN}\\[-1pt]
Source & 9.5&6.7&8.2&29.0&23.4&33.9&27.1&15.9&26.5&47.2&54.7&44.1&30.5&44.5&47.8&29.9 \\
MEMO & 21.6&17.3&20.6&37.1&29.6&40.4&34.4&24.9&34.7&55.1&64.8&54.9&37.4&55.4&57.6&39.1 \\
DDA & 41.3&41.1&40.7&24.4&27.2&30.6&26.9&18.3&27.5&34.6&50.1&32.4&42.3&52.2&52.6&36.1 \\
Tent & 42.2&1.0&43.3&52.4&48.2&55.5&50.5&16.5&16.9&66.4&74.9&64.7&51.6&67.0&64.3&47.7 \\
EATA & 30.1&24.6&34.2&44.3&39.6&48.4&42.4&38.1&46.0&60.7&65.8&61.2&46.7&57.8&59.5&46.6 \\
SAR & 42.7&39.5&41.9&54.6&51.2&58.3&54.4&60.2&54.7&70.3&75.9&66.8&58.4&69.5&66.3&57.6 \\
DeYO & 53.4&50.4&55.0&58.7&59.5&64.5&52.5&68.1&66.3&73.8&78.3&\underline{67.9}&68.9&73.8&70.8&64.1 \\
ReCAP & \underline{53.5}&\underline{56.7}&\underline{56.9}&\underline{59.2}&\underline{60.5}&\underline{65.3}&\textbf{64.0}&\underline{69.6}&\underline{67.2}&\underline{74.1}&\underline{78.4}&64.6&\underline{70.2}&\textbf{74.4}&\underline{71.5}&\underline{65.7} \\
\rowcolor{hdrbg}
\textbf{MASA} & \textbf{54.4}&\textbf{57.2}&\textbf{57.7}&\textbf{60.0}&\textbf{60.9}&\textbf{66.0}&\underline{63.5}&\textbf{70.4}&\textbf{67.5}&\textbf{74.9}&\textbf{78.6}&\textbf{68.9}&\textbf{71.2}&\underline{74.0}&\textbf{72.3}&\textbf{66.5} \\
\bottomrule
\end{tabular}
\end{minipage}\hfill
\begin{minipage}[t]{0.486\textwidth}
\vspace{0pt}
\centering
\caption{Label-shift evaluation on ImageNet-C at severity 5.}
\label{tab:label}
\tiny
\begin{tabular}{@{}l*{16}{c}@{}}
\toprule
  & \multicolumn{3}{c}{Noise}
  & \multicolumn{4}{c}{Blur}
  & \multicolumn{4}{c}{Weather}
  & \multicolumn{4}{c}{Digital}
  & \\
  \cmidrule(lr){2-4}
  \cmidrule(lr){5-8}
  \cmidrule(lr){9-12}
  \cmidrule(lr){13-16}
Method & Gau & Sho & Imp & Def & Gla & Mot & Zoo & Sno & Fro & Fog & Bri & Con & Ela & Pix & JPG & Avg \\
\midrule
\multicolumn{17}{c}{\itshape ResNet50-GN}\\[-1pt]
Source & 17.9&19.9&17.9&19.7&11.3&21.3&24.9&40.4&47.4&33.6&69.2&36.3&18.7&28.4&52.2&30.6 \\
MEMO & 18.4&20.6&18.4&17.1&12.7&21.8&26.9&40.7&46.9&34.8&69.6&36.4&19.2&32.2&53.4&31.3 \\
DDA & \underline{42.5}&43.4&42.3&16.5&19.4&21.9&26.1&35.8&40.2&13.7&61.3&25.2&37.3&46.9&54.3&35.1 \\
Tent & 2.6&3.3&2.7&13.9&7.9&19.5&28.7&16.5&21.9&1.8&70.5&42.2&6.6&49.4&53.7&22.8 \\
EATA & 27.2&28.5&28.4&15.1&16.7&24.6&25.5&32.5&32.2&40.0&66.5&33.2&24.1&42.2&38.6&31.7 \\
SAR & 34.0&36.7&36.2&21.8&20.9&33.2&32.4&38.7&45.6&50.6&\underline{72.9}&46.8&14.3&52.2&56.8&39.5 \\
DeYO & 41.7&44.0&42.5&\underline{23.4}&23.9&\underline{41.3}&13.0&\underline{53.9}&\underline{52.2}&38.6&\textbf{73.1}&52.3&\underline{46.8}&59.3&59.1&44.3 \\
ReCAP & 42.0&\underline{44.1}&\underline{42.7}&19.8&\underline{24.3}&39.7&\textbf{40.2}&46.0&\underline{52.2}&\underline{57.3}&\textbf{73.1}&\underline{52.4}&33.7&\underline{59.4}&\underline{59.5}&\underline{45.8} \\
\rowcolor{hdrbg}
\textbf{MASA} & \textbf{43.3}&\textbf{45.1}&\textbf{43.7}&\textbf{24.2}&\textbf{25.8}&\textbf{42.9}&\underline{37.6}&\textbf{54.5}&\textbf{52.4}&\textbf{59.7}&\textbf{73.1}&\textbf{53.2}&\textbf{47.6}&\textbf{60.3}&\textbf{59.7}&\textbf{48.2} \\
\midrule
\multicolumn{17}{c}{\itshape ViT-Base-LN}\\[-1pt]
Source & 9.4&6.7&8.3&29.1&23.4&34.0&27.0&15.8&26.3&47.4&54.7&43.9&30.5&44.5&47.6&29.9 \\
MEMO & 21.6&17.4&20.6&37.1&29.6&40.6&34.4&25.0&34.8&55.2&65.0&54.9&37.4&55.5&57.7&39.1 \\
DDA & 41.3&41.3&40.6&24.6&27.4&30.7&26.9&18.2&27.7&34.8&50.0&32.3&42.2&52.5&52.7&36.2 \\
Tent & 32.7&1.4&34.6&54.4&52.3&58.2&52.2&7.7&12.0&69.3&76.1&66.1&56.7&69.4&66.4&47.3 \\
EATA & 35.8&34.8&36.8&45.1&47.3&49.3&47.8&56.6&55.5&62.1&72.3&21.6&56.0&64.6&63.7&50.0 \\
SAR & 48.2&\textbf{48.7}&49.0&55.4&54.5&59.2&54.3&55.8&54.5&70.0&76.9&66.1&62.2&70.2&66.5&59.4 \\
DeYO & 53.0&34.4&48.8&\underline{57.6}&58.5&63.3&35.4&67.4&66.0&\underline{73.0}&\underline{77.7}&66.6&68.1&\textbf{73.1}&\underline{69.8}&60.8 \\
ReCAP & \underline{53.1}&38.5&\underline{49.6}&57.3&\underline{59.0}&\underline{63.8}&\underline{60.7}&\underline{67.8}&\underline{66.3}&72.9&\underline{77.7}&\underline{66.8}&\underline{68.2}&\underline{73.0}&\textbf{70.0}&\underline{63.0} \\
\rowcolor{hdrbg}
\textbf{MASA} & \textbf{53.9}&\underline{48.0}&\textbf{50.2}&\textbf{58.5}&\textbf{59.3}&\textbf{64.3}&\textbf{60.9}&\textbf{68.5}&\textbf{66.4}&\textbf{73.5}&\textbf{77.8}&\textbf{67.4}&\textbf{69.0}&72.7&69.6&\textbf{64.0} \\
\bottomrule
\end{tabular}
\end{minipage}
\vspace{-1.0em}
\end{table*}

\begin{table}[!t]
\centering
\captionsetup{font=footnotesize,justification=raggedright,
  singlelinecheck=false,skip=1pt}
\begin{minipage}[t]{0.43\columnwidth}
\vspace{0pt}
\caption{Mixed-domain evaluation on ImageNet-C. We report accuracy (\%) for ResNet50-GN and ViT-Base-LN at corruption severities 5 and 4. Avg averages the results across the two severities.}
\label{tab:mix}
\end{minipage}\hfill
\renewcommand{\arraystretch}{0.92}
\begin{minipage}[t]{0.55\columnwidth}
\vspace{0pt}
\centering
\tiny
\setlength{\tabcolsep}{2.5pt}
\resizebox{\linewidth}{!}{%
\begin{tabular}{@{}lccc@{\hspace{6pt}}lccc@{}}
\toprule
\multicolumn{4}{c}{\itshape ResNet50-GN}
& \multicolumn{4}{c}{\itshape ViT-Base-LN} \\
\cmidrule(lr){1-4}\cmidrule(lr){5-8}
Method & Lv.5 & Lv.4 & Avg
& Method & Lv.5 & Lv.4 & Avg \\
\midrule
Source & 30.6 & 42.7 & 36.7
& Source & 29.9 & 42.9 & 36.4 \\
MEMO & 31.2 & 43.0 & 37.1
& MEMO & 39.1 & 51.3 & 45.2 \\
DDA & 35.1 & 43.6 & 39.4
& DDA & 36.1 & 45.1 & 40.6 \\
Tent & 13.4 & 20.6 & 17.0
& Tent & 16.5 & 64.3 & 40.4 \\
EATA & 38.1 & 47.7 & 42.9
& EATA & 55.7 & 63.7 & 59.7 \\
SAR & 38.3 & 48.6 & 43.5
& SAR & 57.1 & 64.9 & 61.0 \\
DeYO & 38.6 & 50.2 & 44.4
& DeYO & \underline{59.4} & 66.8 & 63.1 \\
ReCAP & \underline{41.5} & \underline{51.2} & \underline{46.4}
& ReCAP & \underline{59.4} & \underline{67.1} & \underline{63.3} \\
\rowcolor{hdrbg}
\textbf{MASA} & \textbf{42.3} & \textbf{52.5} & \textbf{47.4}
& \textbf{MASA} & \textbf{59.8} & \textbf{67.6} & \textbf{63.7} \\
\bottomrule
\end{tabular}
}
\end{minipage}
\vspace{-1.4em}
\end{table}

\noindent\textbf{Experimental setup.}
We evaluate MASA on ImageNet-C \citep{imagenetc} under limited-batch ($=1$), mixed-domain, and time-varying label-shift protocols using ResNet50-GN \citep{resnet} and ViT-Base-LN \citep{vit}. We compare with MEMO \citep{memo}, DDA \citep{dda}, Tent \citep{tent}, EATA \citep{eata}, SAR \citep{sar}, DeYO \citep{deyo}, and ReCAP \citep{recap}. MASA adapts normalization-affine parameters once per batch and uses frozen Qwen2-VL-2B \citep{qwen2vl} and CLIP ViT-B/16 \citep{clip} without a classifier hint. Complete settings are provided in the supplementary material.

\noindent\textbf{Main results.}
Tables~\ref{tab:bs1} and~\ref{tab:label} report corruption-wise accuracy under limited-batch and time-varying label-shift evaluation, respectively. Table~\ref{tab:mix} reports mixed-domain performance at severities 5 and 4 and their average.

\noindent\textbf{Limited-batch adaptation.}
With batch size one, each update lacks within-batch support. MASA obtains 48.4\%/66.5\% accuracy on ResNet/ViT, improving over ReCAP by 2.0/0.8 points. It ranks first on 13 of 15 corruptions for both backbones, showing gains across corruption families. ResNet improves most on elastic transform ($+16.2$) and snow ($+5.8$), but drops by 1.1 points on zoom blur. On ViT, contrast gains 4.3 points, with slight losses on zoom blur and pixelation. This suggests that prototype and semantic evidence is useful when the update target is unreliable.

\noindent\textbf{Time-varying label shift.}
Changing class frequencies can reinforce errors on transiently dominant classes. MASA reaches 48.2\%/64.0\% on ResNet/ViT, exceeding ReCAP by 2.4/1.0 points. On ResNet, it ranks first on 13 corruptions and ties on brightness. Mean gains are positive across all four corruption families (1.1--4.0 points), with elastic transform and snow contributing the most and zoom blur remaining an exception. On ViT, MASA ranks first on 12 corruptions; shot noise improves by 9.5 points over ReCAP but remains 0.7 points below SAR, while pixelation and JPEG compression decrease slightly. The broad gains suggest that reliability filtering helps retain useful memory evidence as the stream prior changes, while the exceptions reveal continued dependence on the current representation.

\noindent\textbf{Mixed-domain adaptation.}
When the corruption changes, earlier evidence can become less relevant. MASA nonetheless leads ReCAP by 1.0/0.4 points on ResNet/ViT, although by smaller margins than in the other protocols. The advantages persist at both severity levels (0.8/1.3 points on ResNet and 0.4/0.5 on ViT at levels 5/4), suggesting that memory remains complementary across heterogeneous shifts.

\noindent\textbf{Feature-space organization.}
Fig.~\ref{fig:feature-space} qualitatively compares the adapted representations. Relative to ReCAP, MASA forms more compact local groups with less intermingling in the highlighted regions, consistent with descriptor-aware retrieval aligning compatible observations with shared prototype targets.

\noindent\textbf{Ablation studies.}
Table~\ref{tab:abl} isolates the main components and loss terms. Removing the prototype loss causes the largest decrease among the tested removals (1.7/0.8 points on ResNet/ViT). Reliability filtering, propagation, and PLPD cause smaller drops of 1.3/0.5, 1.0/0.4, and 0.6/0.2 points. From visual matching at 46.8\%/63.3\%, predictive consistency adds 0.6/0.3 points and semantic agreement adds another 0.8/0.4 points to reach 48.2\%/64.0\%. These monotonic gains support the complementarity of the three cues; the removals further show the value of filtering and propagation before prototype regularization. Additional ablations are provided in the supplementary material.

\begin{table}[t]
\centering
\captionsetup{font=footnotesize,skip=1pt}
\caption{Component ablation and prototype-discrepancy decomposition under label shift (average accuracy, \%).}
\label{tab:abl}
\tiny
\setlength{\tabcolsep}{3.5pt}
\begin{tabular}{lcc@{\hskip 13pt}lcc}
\toprule
Config & RN50 & ViT & Loss terms & RN50 & ViT \\
\midrule
Full & 48.2 & 64.0 & Visual only & 46.8 & 63.3 \\
$-$ proto-loss & 46.5 & 63.2 & $+$ predictive & 47.4 & 63.6 \\
$-$ propagation & 47.2 & 63.6 & $+$ semantic & 48.2 & 64.0 \\
$-$ reliability & 46.9 & 63.5 & \multicolumn{3}{c}{} \\
$-$ PLPD & 47.6 & 63.8 & \multicolumn{3}{c}{} \\
\bottomrule
\end{tabular}
\vspace{-1.8em}
\end{table}

\vspace{-0.5em}
\section{Conclusion}
\vspace{-0.5em}
In this paper, we propose MASA, a WTTA framework designed to address the error reinforcement that can arise when adaptation relies only on evidence from the shifted model. MASA selects sparse, reliability-ranked and visually diverse anchors, obtains structured object and nuisance descriptions from a frozen MLLM, and propagates these descriptors through local feature neighborhoods. We further consolidate visual, predictive, and semantic evidence in a bounded prototype memory, where descriptor-aware retrieval supplies auxiliary targets for normalization-affine adaptation without treating MLLM outputs as class labels. Experiments across limited-batch, mixed-domain, and time-varying label-shift protocols demonstrate consistent improvements with both ResNet and ViT, while the ablations support the complementary roles of prototype consistency, propagation, reliability filtering, and semantic agreement. We hope this work encourages WTTA research to move beyond model-internal signals and explore sparse semantic grounding for online adaptation.

\bibliography{gta_arxiv}

\clearpage
\appendix
\setcounter{secnumdepth}{2}
\twocolumn[
\begin{center}
{\Large\bfseries Sparse MLLM Anchors, Dense Adaptation:\\
Breaking the Self-Referential Loop in Wild Test-Time Adaptation\par}
\vspace{0.4em}
{\large\bfseries Appendix\par}
\end{center}
\vspace{0.5em}
]

\noindent\textbf{Appendix organization.}
Appendix~A reviews closely related work. Appendix~B gives the complete online procedure, Appendix~C specifies the implementation, Appendix~D presents additional ablations and analyses, and Appendix~E discusses limitations.

\section{Related Work}

\subsection{Wild Test-time Adaptation}
TTA updates a pretrained model using unlabeled test data to address distribution shifts encountered during inference \citep{tent,lame,cotta,adacontrast,ecotta}. WTTA focuses on online conditions such as single-sample batches, mixed-domain streams, and time-varying class imbalance \citep{sar,recap}. Methods designed for these conditions stabilize adaptation in several ways. Sample-selection approaches filter updates using entropy, sharpness, or perturbation sensitivity \citep{eata,sar,deyo}. Memory- and prototype-based methods accumulate structure from the stream \citep{t3a,nsp,adaneg}, while ReCAP \citep{recap} replaces pointwise entropy with confidence and consistency over local feature regions. Despite these differences, the evidence used for adaptation is derived predominantly from the predictions or representations of the model being adapted. Input restoration \citep{dda} and diffusion-score distillation \citep{dusa} provide additional priors, but their guidance remains implicit at the pixel or score level. MASA instead uses explicit descriptions of image content and nuisance factors as complementary adaptation evidence.

\subsection{MLLMs as External Semantic Knowledge}
Vision--language models connect recognition with semantic concepts expressed in text. OOD methods use text to represent negative concepts, outliers, or visually confusing classes, and to refine the scoring of image--text evidence \citep{eoe,neglabel,mcm,clipn,locoop,negprompt,idlike,lapt,oodd}. ANTS \citep{ants} further uses descriptions from a frozen MLLM to update the negative textual space of a frozen, CLIP-native detector. MLLMs have also served as pseudo-label generators in offline source-free adaptation \citep{rcl}. MASA assigns a different role to language outputs. It neither places them in the classifier's decision space nor treats them as class labels. Instead, sparse MLLM descriptions characterize selected test samples and provide persistent guidance for online adaptation of a closed-set classifier that need not be text aligned.

\section{Complete Online Procedure}
\begin{algorithm}[H]
\caption{Online MASA at test step $t$}
\label{alg:gta}
\small
\begin{algorithmic}[1]
\REQUIRE Batch $X_t$; $\theta_t$; incoming states and auxiliary histories
\STATE Compute pre-update $\mathbf{o}_i$, $\mathbf{p}_i$, $\mathbf{z}_i$, $\mathbf{v}_i$, $\ell_{\mathrm{RE},i}$, and $\ell_{\mathrm{RI},i}$
\STATE Compute $m_i$, $\phi_i$, and $q_i$; append $\{\mathsf{r}_{t,i}\}_{i=1}^{B_t}$ to form $\widetilde{\mathcal{W}}_t$
\IF{semantic refresh is triggered}
  \STATE Rank by Eq.~\eqref{eq:anchor-ranking}; diversify by farthest-point sampling
  \STATE Query and encode anchors; append to $\mathcal{A}_t$ and immediately upsert $\mathcal{M}_t$
\ENDIF
\STATE Select by Eq.~\eqref{eq:region-selection}; compute $\mathcal{L}_{\mathrm{region}}$
\STATE Propagate, retrieve, and compute $\mathcal{L}_{\mathrm{proto}}$
\STATE Take one step on Eq.~\eqref{eq:masa-objective} to obtain $\vartheta_{t+1}$
\STATE Sequentially write the batch; then maintain and age memory
\STATE Update $\bar\ell_t$ by Eq.~\eqref{eq:recovery}; recover if $\bar\ell_t<\tau_{\mathrm{rec}}$
\RETURN Pre-update logits $\mathbf{o}_i$
\end{algorithmic}
\end{algorithm}

\begin{figure*}[t]
\centering
\begin{tikzpicture}
\begin{groupplot}[
  group style={group size=3 by 1,horizontal sep=0.75cm},
  width=0.29\textwidth,
  height=0.205\textwidth,
  ymin=45.5,
  ymax=49.0,
  ytick={46,46.5,47,47.5,48,48.5},
  ymajorgrids,
  grid style={black!12},
  axis line style={black!65},
  tick align=outside,
  tick label style={font=\scriptsize},
  label style={font=\small},
  title style={font=\small\bfseries},
  legend columns=3,
  legend style={
    draw=none,
    font=\footnotesize,
    /tikz/every even column/.append style={column sep=0.9em}
  }
]
\nextgroupplot[
  title={(a) Prototype weight},
  xlabel={$\lambda_{\mathrm{proto}}$},
  ylabel={Accuracy (\%)},
  legend to name=hparamlegend,
  xmin=0.05,
  xmax=0.85,
  xtick={0.1,0.2,0.4,0.8}
]
\addplot[
  curveblue,thick,mark=*,mark size=2.2pt,mark options={fill=white},
  point meta=y,
  nodes near coords={\pgfmathprintnumber[fixed,fixed zerofill,precision=1]{\pgfplotspointmeta}},
  every node near coord/.append style={
    font=\scriptsize,text=black,anchor=south,yshift=1.5pt,inner sep=0.5pt
  }
]
  coordinates {(0.1,47.9) (0.2,48.2) (0.4,48.0) (0.8,47.5)};
\addlegendentry{MASA (ResNet50-GN)}
\addplot[baselinegray,dashed,thick,domain=0.05:0.85,samples=2] {45.8};
\addlegendentry{ReCAP}
\addplot[defaultred,only marks,mark=square*,mark size=3pt]
  coordinates {(0.2,48.2)};
\addlegendentry{Default}

\nextgroupplot[
  title={(b) Refresh interval},
  xlabel={$T_{\mathrm{ref}}$},
  xmin=8,
  xmax=270,
  xtick={16,64,128,256},
  yticklabels=\empty
]
\addplot[
  curveblue,thick,mark=*,mark size=2.2pt,mark options={fill=white},
  point meta=y,
  nodes near coords={\pgfmathprintnumber[fixed,fixed zerofill,precision=1]{\pgfplotspointmeta}},
  every node near coord/.append style={
    font=\scriptsize,text=black,anchor=south,yshift=1.5pt,inner sep=0.5pt
  }
]
  coordinates {(16,48.3) (64,48.2) (128,47.8) (256,47.6)};
\addplot[baselinegray,dashed,thick,domain=8:270,samples=2] {45.8};
\addplot[defaultred,only marks,mark=square*,mark size=3pt]
  coordinates {(64,48.2)};

\nextgroupplot[
  title={(c) Memory capacity},
  xlabel={$K_{\max}$},
  xmin=20,
  xmax=270,
  xtick={32,64,128,256},
  yticklabels=\empty
]
\addplot[
  curveblue,thick,mark=*,mark size=2.2pt,mark options={fill=white},
  point meta=y,
  nodes near coords={\pgfmathprintnumber[fixed,fixed zerofill,precision=1]{\pgfplotspointmeta}},
  every node near coord/.append style={
    font=\scriptsize,text=black,anchor=south,yshift=1.5pt,inner sep=0.5pt
  }
]
  coordinates {(32,47.7) (64,48.2) (128,48.1) (256,47.9)};
\addplot[baselinegray,dashed,thick,domain=20:270,samples=2] {45.8};
\addplot[defaultred,only marks,mark=square*,mark size=3pt]
  coordinates {(64,48.2)};
\end{groupplot}
\node[anchor=south] at ([yshift=30pt]group c2r1.north)
  {\pgfplotslegendfromname{hparamlegend}};
\end{tikzpicture}
\caption{\textbf{Hyperparameter sensitivity under label shift.}
Average accuracy (\%) with ResNet50-GN when varying
(a) the prototype-loss weight, (b) the periodic refresh interval, and
(c) the memory capacity. Each sweep changes one parameter while fixing
the others. Red squares mark the default settings, and the dashed line
shows the ReCAP baseline under the same protocol.}
\label{fig:hparam}
\end{figure*}
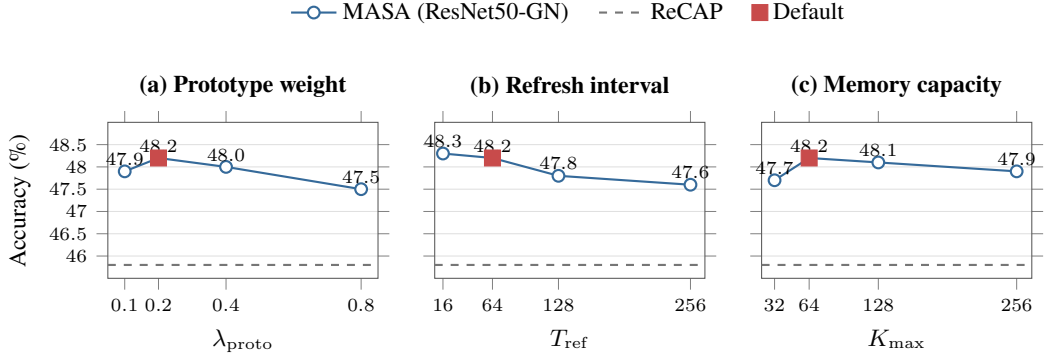

\section{Implementation Details}
We use the released WTTA implementation with ResNet50-GN \citep{resnet} and ViT-Base-LN \citep{vit} from \texttt{timm}. Optimization uses SGD with momentum $0.9$, base learning rates $0.00025/0.001$ for ResNet/ViT with the released batch-size adjustment, and one normalization-affine update per batch. We optimize BN/LN/GN affine parameters while excluding ResNet layer~4, ViT blocks~9--11, and the final normalization layer. The normalized memory feature is extracted by an additional forward pass before the update.

Following ReCAP \citep{recap}, the regional path loads the fixed, backbone-specific diagonal feature statistic distributed with its implementation. It does not sample or retain a calibration subset. We use $\rho_{\mathrm{reg}}=12$, $\lambda_{\delta}=5\times10^{-4}$, $\lambda_{\mathrm{RI}}=0.5$, $P=4$, and $\tau_{\mathrm{PLPD}}=0.2$. For ResNet, $\tau_{\mathrm{RE}}=0.8\log C$ and $\omega_{\max}^{\mathrm{reg}}=3$. For ViT, the corresponding values are $\log C$ and $1.5$. In the batch-size-one protocol, the cap is set to $\omega_{\max}^{\mathrm{reg}}=5$. The reliability gate uses $H_{\mathrm{hist}}=32$ and $(\tau_{\mathrm{RI}}^{\mathrm{anc}},\tau_m,\tau_\phi)=(10,0.05,0.5)$. Anchor selection uses $(c_{\mathrm{clip}},B_a,r_{\mathrm{pool}})=(2,2,8)$.

Unless otherwise stated, MASA uses frozen Qwen2-VL-2B \citep{qwen2vl} and CLIP ViT-B/16 \citep{clip}, with no classifier hint. We set $(\kappa_{\min},K_a)=(0.1,4)$ and $(\tau_a,\epsilon_{\mathcal C})=(0.07,0.15)$ for semantic grounding and propagation. Prototype retrieval uses $(\tau_q,\tau_{\mathrm{assign}})=(0.1,0.6)$ and $(\alpha_{\mathrm{age}},\alpha_{\mathrm{cand}})=(0.02,0.05)$, with $\tau_\omega=0.1$ for prototype weighting. We set $\lambda_{\mathrm{proto}}=0.2$ and $(\rho_{\mathrm{rec}},\tau_{\mathrm{rec}})=(0.9,0.4)$ for the objective and recovery. Memory updates use $(\eta_0,\tau_{\mathrm{store}},\lambda_{\mathrm{stat}})=(0.25,0.5,0.05)$ and $n_{\min}=2$. Finally, $K_{\max}=64$, $(H_D,H_{\mathrm{win}},H_{\mathrm{cov}},T_{\mathrm{ref}})=(8,128,128,64)$, and $(\tau_D,\tau_{\mathrm{cov}})=(0.45,0.25)$. Baseline results in Tables~\ref{tab:bs1}, \ref{tab:label}, and~\ref{tab:mix} are taken from ReCAP \citep{recap} under the same benchmark.

\section{Additional Ablations and Analyses}
Beyond the component removals in Table~\ref{tab:abl}, we examine where the semantic gains originate and how MASA behaves under its main control parameters.

\subsection{Semantic Source and Class Information}
We vary the semantic source while keeping the remaining components and hyperparameters fixed (Table~\ref{tab:mllm}). The comparison ranges from no semantic input and a deterministic text-hash control to BLIP-2 \citep{blip2}, LLaVA-1.5-7B \citep{llava}, and Qwen2-VL-2B \citep{qwen2vl}. The hash control maps descriptor tokens to fixed vectors without a learned text encoder. Table~\ref{tab:hint} separately measures the effect of providing class hints to the MLLM, including a ground-truth oracle that is not used by MASA.

\begin{table}[H]
\centering
\caption{Effect of the semantic source on average accuracy (\%) under each WTTA protocol. Here, bs1, mix, and lbl denote limited-batch, mixed-domain, and label-shift evaluation.}
\label{tab:mllm}
\small
\setlength{\tabcolsep}{3.5pt}
\begin{tabular}{lcccccc}
\toprule
Semantic source & \multicolumn{3}{c}{ResNet50-GN} & \multicolumn{3}{c}{ViT-Base-LN}\\
 & bs1 & mix & lbl & bs1 & mix & lbl \\
\midrule
None & 47.0 & 46.7 & 47.4 & 65.9 & 63.3 & 63.6 \\
Hash (text) & 47.2 & 46.8 & 47.3 & 66.0 & 63.2 & 63.5 \\
BLIP-2 & 47.7 & 47.0 & 47.7 & 66.2 & 63.4 & 63.7 \\
LLaVA-1.5-7B & 48.1 & 47.2 & 47.9 & 66.3 & 63.5 & 63.8 \\
Qwen2-VL-2B & 48.4 & 47.4 & 48.2 & 66.5 & 63.7 & 64.0 \\
\bottomrule
\end{tabular}
\end{table}

\paragraph{Effect of the semantic source.}
The None and Hash controls remain close under every protocol: hashing changes accuracy by at most 0.2 points and slightly reduces label-shift performance on both backbones. Token identity alone therefore provides little useful structure for prototype retrieval. Learned semantic sources give a different pattern. Performance increases consistently from BLIP-2 to LLaVA-1.5-7B and Qwen2-VL-2B across all six backbone--protocol combinations. Relative to using no semantics, Qwen2-VL-2B improves limited-batch, mixed-domain, and label-shift accuracy by 1.4/0.7/0.8 points on ResNet and 0.6/0.4/0.4 points on ViT. The larger ResNet gain, particularly with batch size one, may reflect both its lower starting accuracy and the absence of support from other samples in the current batch. The smaller but consistent ViT gains show that semantic evidence remains complementary when the visual representation is already stronger. Overall, the ordering is consistent with more informative descriptions producing better semantic matches, rather than the benefit arising from an arbitrary extra code.

\begin{table}[H]
\centering
\caption{Effect of class information in the MLLM prompt. Results are average accuracy (\%) under label shift. MASA uses the no-hint setting.}
\label{tab:hint}
\small
\setlength{\tabcolsep}{7pt}
\begin{tabular}{lcc}
\toprule
Hint to MLLM & RN50 & ViT \\
\midrule
None & 48.2 & 64.0 \\
Predicted class & 47.9 & 63.8 \\
Ground-truth class (oracle) & 49.1 & 64.6 \\
\bottomrule
\end{tabular}
\end{table}

\paragraph{Effect of class information.}
The no-hint setting reaches 48.2\%/64.0\% on ResNet/ViT, whereas conditioning the MLLM on the classifier's predicted class lowers accuracy by 0.3/0.2 points. A predicted hint can steer the description toward the model's current belief, including an incorrect one, and thereby partially reintroduce the feedback that external semantics is intended to reduce. In contrast, the ground-truth oracle reaches 49.1\%/64.6\%, showing that correct category information can make the resulting descriptors more discriminative. The gap between predicted and oracle hints emphasizes that the value of class information depends on its correctness. MASA therefore uses no class hint: this retains an image-only semantic query while avoiding unavailable oracle information.

\subsection{Hyperparameter Sensitivity}
Figure~\ref{fig:hparam} examines three parameters with distinct roles: the prototype-loss weight $\lambda_{\mathrm{proto}}$, the periodic refresh interval $T_{\mathrm{ref}}$, and the memory budget $K_{\max}$. These sweeps characterize the balance between regional and prototype adaptation, the accuracy--query-cost trade-off, and sensitivity to memory capacity.

\paragraph{Sensitivity trends.}
The prototype weight has a broad optimum around the default $\lambda_{\mathrm{proto}}=0.2$. Reducing it to 0.1 or increasing it to 0.4 changes accuracy by only 0.3 and 0.2 points, respectively, whereas a larger value of 0.8 reduces accuracy to 47.5\%. This decline is consistent with prototype consistency dominating the regional objective and over-constraining updates toward imperfect or stale memory entries. Refreshing every 16 steps gives 48.3\%, only 0.1 points above the default interval of 64 but with four times as many periodic refresh opportunities. Longer intervals of 128 and 256 reduce accuracy to 47.8\% and 47.6\%, suggesting that descriptors become less representative as the stream evolves. Memory capacity shows a similar saturation pattern: increasing $K_{\max}$ from 32 to 64 improves accuracy from 47.7\% to 48.2\%, while capacities of 128 and 256 yield 48.1\% and 47.9\%. A small memory cannot cover enough recurring structure, whereas a much larger one may preserve redundant or outdated clusters. The defaults therefore lie near the stable region of each sweep while balancing adaptation strength, refresh frequency, and memory capacity.

\FloatBarrier

\newpage
\section{Limitations}
First, our evaluation focuses on ImageNet-C classification. Synthetic corruptions provide controlled WTTA protocols, but they do not cover natural domain drift, open-set arrivals, or dense prediction tasks, where the usefulness and granularity of semantic descriptors may differ. Evaluating MASA on these settings is necessary to establish how broadly sparse semantic grounding transfers beyond image classification.
Second, MASA requires a frozen MLLM and a text encoder. Sparse anchor queries amortize their cost across neighboring samples, but do not remove the added latency and memory footprint, which also depend on hardware and model-serving choices. Incorrect or underspecified descriptions may still enter the prototype memory despite reliability filtering. Lightweight semantic encoders, adaptive query budgets, and explicit descriptor uncertainty are promising directions.
Third, external descriptions do not make the complete adaptation path independent of the classifier. Anchor ranking, propagation neighborhoods, and transition statistics still use its predictions and feature geometry. If the representation deteriorates severely or class identities alternate faster than the history window can track, these operations may select or propagate unreliable evidence. A more complete treatment should jointly model semantic uncertainty and rapid stream dynamics.

\end{document}